\documentclass[review]{elsarticle}

\usepackage{lineno,hyperref}
\modulolinenumbers[5]

\usepackage{amsmath}
\usepackage{amsfonts}
\usepackage{amssymb}
\usepackage{graphicx}
\usepackage{multirow,multicol}
\usepackage{pifont,hyperref,lastpage,fancyhdr,movie15,float}
\usepackage{eqnarray}
\usepackage{geometry}
\usepackage{caption}
\usepackage{subcaption}
\usepackage{pdfpages}
\usepackage{tikz}
\usepackage{color}
\usepackage{listings}
\usepackage{color, colortbl}
\usepackage{csquotes}
\usepackage{algorithm2e}

\DeclareMathOperator*{\argmin}{arg\,min}

\begin{document}

\begin{frontmatter}

\title{Label Assisted Autoencoder for Anomaly Detection in Power Generation Plants}
\author[mymainaddress]{Marcellin Atemkeng \corref{mycorrespondingauthor1}}
\author[mysecondaryaddress]{Victor Osanyindoro}
\cortext[mycorrespondingauthor1]{m.atemkeng@ru.ac.za}
\author[mysecondaryaddress]{Rockefeller Rockefeller}
\author[mymainaddress,mythirdaddress]{Sisipho Hamlomo}
\author[mysecondaryaddress]{Jecinta Mulongo}
\author[myfourthaddress]{Theophilus Ansah-Narh}
\author[myfifthaddress,mymainaddress]{Franklin Tchakounte}
\author[mysixthaddress]{Arnaud Nguembang Fadja}

\address[mymainaddress]{Department of Mathematics, Rhodes University, 6139 Makhanda, South Africa}
\address[mysecondaryaddress]{African Institute for Mathematical Sciences, Limbe, Cameroon}
\address[mythirdaddress]{Department of Statistics, Rhodes University, 6139 Makhanda, South Africa}
\address[myfourthaddress] {Ghana Space Science and Technology Institute, Ghana Atomic Energy Commission, Accra, Ghana}
\address[myfifthaddress] {Department of SFTI, School of Chemical Engineering and Mineral Industries, University of Ngaoundéré, Cameroon}
\address[mysixthaddress]{Dipartimento di Ingegneria, University of Ferrara, Via Saragat 1, 44122 Ferrara, Italy}
\begin{abstract}
One of the critical factors that drive the economic development of a country and guarantee the sustainability of its industries is the constant availability of electricity. This is usually provided by the national electric grid. However, in developing countries where companies are emerging on a constant basis including telecommunication industries, those are still experiencing a non-stable electricity supply. Therefore, they have to rely on generators to guarantee their full functionality. Those generators depend on fuel to function and the rate of consumption gets usually high, if not monitored properly. Monitoring operation is usually carried out by a (non-expert) human. In some cases, this could be a tedious process, as some companies have reported an exaggerated high consumption rate. This work proposes a label assisted autoencoder for anomaly detection in the fuel consumed by power generating plants. In addition to the autoencoder model, we added a labelling assistance module that checks if an observation is labelled,  the label is used to check the veracity of the corresponding anomaly classification given a threshold. A consensus is then reached on whether
training should stop or whether the threshold should be updated or the training should continue with the search for hyper-parameters. 
Results show that the proposed model is highly efficient for reading anomalies with a detection accuracy of 97.20\% which outperforms the existing model of 96.1\% accuracy trained on the same dataset. In addition, the proposed model is able to classify the anomalies according to their degree of severity.
\end{abstract}


\end{frontmatter}


\section{Introduction}
About 3\% of the world’s electrical energy is utilised by the information communication technology  companies \citep{humar2011rethinking} and  the telecommunication industry is one of the fastest growing industrial sectors among Agriculture, Banking, Infrastructure, and Oil and Gas. The number of telecommunication industries and the quest for expansion and growth has led to an increase in base stations across targeted countries to boost their network coverage and enhance the effective flow of communication. With the increase in the number of base stations, the issue of base station management needs to be addressed. 
Grid energy has been known to be the main source of  power in developing countries such as those in Africa, and it is expected that these base stations located across different rural and urban areas will be powered by grid energy. However, electricity is quite unstable in most parts of these developing countries and this has forced base stations to look for a reliable alternative source of energy. These alternatives include photovoltaic panels (PV), wind turbines and diesel generators, but mostly generators due to a lack of space for the installation of PV or wind turbines \cite{lorincz2013renewable}. The high cost of fuel and its transportation to supply stations located in  rural areas has increased the operational cost of these companies.
These generators are being refilled manually,  thus creating room for irregular or unusual fuel consumption which might be caused by several reasons such as fuel theft, fuel leakages or poor maintenance of equipment. A study conducted in a base station in Cameroon has shown that the design of the base station building, room cooling systems such as air conditioners and careless handling of lights increase the rate of fuel consumption in the base station \cite{ayang2016power}. Espadafor et al. \cite{espadafor2009analysis} also attested that generator performance can be affected by the age of the generator, the number of loads powered by the generator and improper maintenance. 

In Cameroon,  TeleInfra  LTD is one such company whose objective is to manage base stations in various parts of the country. The services include maintenance of base stations and refuelling of generators. Like any other businesses relying on grid energy, the unstable power supply has resulted in high operating costs as companies have to find alternative sources of power supply to sustain the continuous operations of the business. The use of alternative sources such as solar panels, hybrid energy and generators has been implemented by  TeleInfra for the sustainability of the business performance. Data on fuel consumed such as working hours of the generator, the quantity of fuel refuelled, the rate of consumption, generator maintenance, and total fuel consumed, are collected from base stations \cite{mulongo2020anomaly}. 

Although many strategies have been proposed to improve energy saving such as building a well-ventilated base station, the use  of air conditioners as a cooling system and  heat pipes to remove hot air from the base station \cite{jinggang2009feasibility}. However, detecting such irregularities is challenging especially when there are numerous base stations functioning at once. 
Anomaly detection in power generating plants is aimed at detecting such irregularities in the behaviour of the data provided. Although there are different algorithms for detecting anomalies, machine learning algorithms are the most used and popular for anomaly detection due to their ability for automation and their effectiveness in the context of deep learning, especially when involving large datasets. 
According to Goodfellow \cite{goodfellow2016deep}, deep learning is a variant of applied statistics with an increasing emphasis on using computers in estimating complex functions statistically, but a reduced emphasis on proving confidence intervals around these functions. Machine learning algorithms come in several variants. In supervised learning, models can make predictions on unlabelled data after they have been trained on labelled data whereas, in unsupervised learning, models can only make predictions on unlabelled data by learning similar features and patterns embedded in the datasets. In reinforcement learning, a goal is given and an agent undergoes training in an environment with the purpose of finding an optimal solution to accomplishing the goal.

Anomalies can be an indicator of areas that require attention, and detecting them has been quite popular among the research community. In the past, the research community has conducted several anomaly detections ranging from comprehensive to certain application domains with different machine learning algorithms. Mulongo et al. \cite{mulongo2020anomaly} worked on a similar area, four different supervised learning algorithms were used in their work for detecting an anomaly in a power generation plant.  However, in real-life scenarios, abnormal behavioural patterns are very few compared to normal behaviour. For instance, in the same work of Mulongo et al. \cite{mulongo2020anomaly}, labels were created by the authors based on certain criteria and only 35 per cent of the dataset turned out to be anomalous and the authors had to duplicate the data to balance between the two categories.  These are key challenges in recognising anomalies with a supervised learning algorithm since they greatly rely on labelling and balance between normal and anomalous data patterns. The goal is to learn how efficiently a generating plant should behave so that observations that do not confine to the norm will be identified, and the required attention can be given. Our work investigates an alternative approach for anomaly detection in a power generation plant in an unsupervised manner based on the dataset used in Mulongo et al. \cite{mulongo2020anomaly}. The aim is to compare performance with the results  obtained in Mulongo et al. \cite{mulongo2020anomaly}. The proposed unsupervised learning framework is built from a modifying and fine-tuning  autoencoder  which is free of the hassle of data labelling and balancing. An additional module is added to an autoencoder, the new module uses some labelled data to check if each of the observations is correctly classified and then one of the three steps below is activated:
\begin{itemize}
    \item Update the  threshold of the autoencoder which therefore increases the model overall accuracy to an acceptable  level
    \item Update the interval of variation for numerical hyper-parameters, the best values of the hyper-parameters are then  explored in the new search space
    \item Provide overall performance score
\end{itemize}
This paper is organised as follows: Section~\ref{chap:relatedwork} explains anomaly detection and investigates related works for anomaly detection in power grid plants. Section~\ref{chap:autometrics} discusses autoencoders  which replicate the input data through a compressed representation. This section  also discusses the different evaluation metrics used in this work. Section~\ref{chap:proposed} proposes the label assisted autoencoder and provides a detail discussion.  The dataset and feature engineering are discussed in Section~\ref{chap:dataset}. Section~\ref{chap:results} discusses the results and limitations, and  Section~\ref{chap:conclud} concludes the work.

\section{Anomaly detection and related works}
\label{chap:relatedwork}
Anomalies also known as outliers often refer to instances or data samples that are significantly distanced from the main body of an examined data \cite{aggarwal2017introduction}. These distanced values often
indicate a deviation from its established normal pattern which can sometimes be a measurement error or an indication of a data sample of a different population~\cite{demestichas2021comparative}.  Outliers classification depends on the type and  domain of the given data as well as the data analyst. Since many outliers 
are linked directly with abnormal behaviour, they are also referred to as deviants, anomalies, or abnormalities in the literature of statistics and data analysis \cite{aggarwal2017introduction}. 
According to \cite{aggarwal2017introduction}, interpreting data is directly associated with the detection of  anomalous samples. Demestichas et al. \cite{demestichas2021comparative} suggested that it is essential to achieve the highest possible interpretability level to properly select the best anomaly detection method from different ranges of  the relevant algorithm. 
There are two major categories of anomalies depending on the given dataset; multivariate and univariate \cite{demestichas2021comparative}. Multivariate anomalies can be spotted in multi-dimensional data while univariate anomalies are spotted in a single-dimensional data. Besides the two categories of anomaly, there are other categories which depend on
the distribution of the given data.  Data samples that are considered anomalous when viewed against the entire  dataset are point anomalies, while  data samples that are considered anomalous with respect to meta-information related to the data sample are contextual anomalies \cite{hayes2015contextual}. In other words, contextual anomalies are classified based on local neighborhoods, while point anomalies are classified based on the overall dataset. Collective anomalies denote anomalous data collection
samples which together are considered an anomalous pattern. 

Fahim and Sillitti \cite{fahim2019anomaly}  provide two anomaly detection methods;  statistical and machine learning methods. The statistical method uses various algorithms such as density-based, distance-based, parametric and statistical-based. However, Trinh et al. \cite{trinh2019detecting} noted that one of the major challenges that are encountered by this approach is the design of a suitable model that can accurately separate normal data from unusual data points. On the other hand, machine learning methods consist of both supervised and unsupervised learning algorithms in which dataset can either be labelled for supervised learning or unlabelled for unsupervised learning. Some advantages of this method are an enhancement of detection speeds and its ability to handle complexity with  the less human intervention~\cite{omar2013machine}.

Many researchers have worked on different machine learning techniques for anomaly detection, but most of the current works applied artificial neural networks (ANNs) to classification tasks. The labelled data is used during the training stage, and then the learned model  is able to correctly classify sample data never used during the training process. This technique is generally classified under supervised machine learning techniques. Such an example is trained in Mulongo et al \cite{mulongo2020anomaly} in which support vector machines (SVM \cite{vishwanathan2002ssvm}), K-Nearest Neighbours (KNN \cite{peterson2009k}), Logistic Regression
(LR \cite{nick2007logistic}), and  MultiLayer Perceptron (MLP \cite{ruck1990feature}) are used for anomaly detection associated with the fuel consumed dataset from an energy company. However, the energy sector is not the only place anomaly detection with supervised machine learning has been applied, others include fraud detection in credit card \cite{bhattacharyya2011data} attacks and anomaly detection in IoT sensors \cite{hasan2019attack}.
One of the main advantages of  supervised learning techniques is the ability to handle high-dimensional datasets with high-performance \cite{said2020network}. However, there is a major problem with this technique. When dealing with real life data, the majority of them contains fewer anomalies data which is quite challenging and can cause the problem of an unbalanced dataset. This is an issue for  supervised learning techniques since they greatly rely on labelled and balanced data.
However, unsupervised learning techniques can be used to address this problem, for example, the autoencoder considers a specific kind of feed-forward neural network that can be applied in outlier-based anomaly detection rather than classification problems. Hawkins et al. \cite{hawkins2002outlier} proposed an approach that involved autoencoder for outlier detection, however many researchers have investigated hybrid  methods; e.g. \cite{said2020network} proposed an approach based on long short term memory (LSTM) autoencoder and One-class SVM (OC-SVM). The approach is used to detect anomalies based attacks in an unbalanced dataset. The idea is to use the LSTM-autoencoder to train a model to learn the pattern in the normal class (dataset without anomaly) so that-the model is able to replicate the input data at the output layer with a small reconstruction error.  When there are anomalies in the data, the model fails to replicate the anomalous samples. This arises when the reconstruct error is very high.

Another unsupervised learning technique is the k-means, Zhang et al. \cite{zhang2021power}, used the transformer model and the k-means clustering method for anomaly detection. The k-means was also used in the work of Munz et al. \cite{munz2007traffic}, to detect traffic anomalies, the main idea is to train data containing unlabelled records and separate them into clusters of normal and anomalous data.

\section{Autoencoder and performance metrics}
\label{chap:autometrics}
\subsection{Autoencoder}
\begin{figure}
	\centering
	\includegraphics[width=1\textwidth]{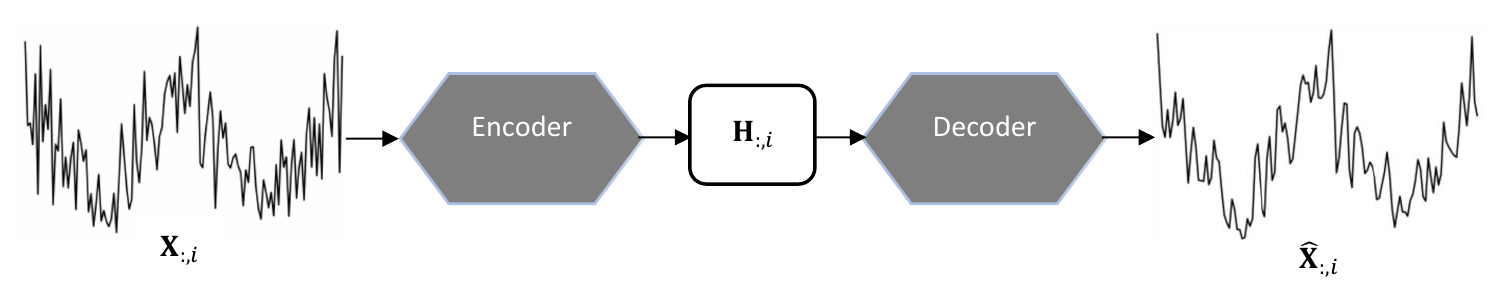}
	\caption{Simple architecture of an autoencoder (adapted from \cite{bank2020autoencoders}).}
	\label{fig:auto}
\end{figure}

An autoencoder is a neural network that is trained to attempt to copy its input to its output in an unsupervised manner. It consists of three parts: An Encoder part, a code and a decoder part. The encoder compresses the input data $\mathbf{X}$ defined in two dimensions as:
\begin{align}
     \mathbf{X} &=\begin{pmatrix}
   \mathbf{X}_{1,1} & \mathbf{X}_{1,2} & \cdots & \mathbf{X}_{1,M} \\
   \mathbf{X}_{2,1} & \mathbf{X}_{2,2} & \cdots & \mathbf{X}_{2,M} \\
   \vdots  & \vdots  & \ddots & \vdots  \\
   \mathbf{X}_{N,1} & \mathbf{X}_{N,2} & \cdots & \mathbf{X}_{N,M} 
 \end{pmatrix}\in \mathbb{R}^{N\times M}
\end{align}
where $\mathbf{X}_{:,i}\in \mathbb{R}^{N}$ is an observation with $N\in \mathbb{R}$ entries among the $M\in \mathbb{R}$ observations. The encoder compressed $\mathbf{X}$   to a lower-dimensional space $ \mathbf{H}_{:,i} \in\mathbb{R}^Q$. The decoder produces a predicted output $ \hat{\mathbf{X}}$,
\begin{align}
     \hat{\mathbf{X}} &=\begin{pmatrix}
    \hat{\mathbf{X}}_{1,1} &  \hat{\mathbf{X}}_{1,2} & \cdots &  \hat{\mathbf{X}}_{1,M} \\
    \hat{\mathbf{X}}_{2,1} &  \hat{\mathbf{X}}_{2,2} & \cdots &  \hat{\mathbf{X}}_{2,M} \\
   \vdots  & \vdots  & \ddots & \vdots  \\
    \hat{\mathbf{X}}_{N,1} &  \hat{\mathbf{X}}_{N,2} & \cdots &  \hat{\mathbf{X}}_{N,M} 
 \end{pmatrix}\in \mathbb{R}^{N\times M}
\end{align}
by reconstructing the original data from the compressed representation. 
It makes use of multiple layers and uses non-linear activation functions to learn the non-linear  relationship embedded in the data. The goal is to make the reconstruction error as minimal as possible, which  means finding the parameters that make the reconstruction $\hat{\mathbf{X}}$ as close as possible to the original input $\mathbf{X}$. 
Autoencoder can be applied to various tasks such as anomaly detection \cite{said2020network,zhou2017anomaly},  generative model \cite{mishra2018generative,johnson2017generative}, clustering \cite{ge2019dual,guo2017deep}, classification \cite{luo2017convolutional}, recommendation systems \cite{ferreira2020recommendation} and dimensionality reduction \cite{sakurada2014anomaly,wang2014generalized}. Figure \ref{fig:auto} depicts a simple architecture of an autoencoder for which the detailed description is as follows:

	 \textit{Encoding:} During the encoding process, assuming  the input data $ \mathbf{X}_{:,i} $ is a high dimensional vector that is mapped to a low dimensional vector $ \mathbf{H}_{:,i} $ after filtering 
	  insignificant features. This is expressed mathematically  as:
	\begin{equation}\label{eqn:3.1.1a}
		\mathbf{H}_{:,i} = f(\mathbf{X}_{:,i}),
	\end{equation}
	where $ f $ is a neural network that is trained with sets of activation functions weights and biases. Note that $f \in \mathbb{R}^{N} \rightarrow \mathbb{R}^{Q}$, where $Q$ is the number of compressed representations.
	
 \textit{Decoding:} During the decoding process, the compressed representation $ \mathbf{H}_{:,i}$ of $\mathbf{X}_{:,i}$ is used to generate the output $\hat{\mathbf{X}}_{:,i}$ that maps back into the reconstruction of $ \mathbf{X}_{:,i} $:
	\begin{align}\label{eqn:3.1.1}
		\hat{\mathbf{X}}_{:,i} &= g (\mathbf{H}_{:,i})\\
		        &=g\big(f(\mathbf{X}_{:,i})\big),
	\end{align}
where $g$ is the decoding neural network with activation functions, weights, and biases that could be completely independent of the corresponding activation, weights, and biases of the encoding neural network. As discussed in \citep{michelucci2022introduction, wei2023lstm}, the autoencoder involves the search for $f$ and $g$ which minimizes the average of the loss $\Delta_{\theta, \phi}$ between all the samples $\mathbf{X}_{:,i}$, added for example, to a $l_2$  weighted regularization term $a_j$:
\begin{align}
    [\hat{f}, \hat{g}]&=\argmin_{f, g}\bigg(<\Delta_{\theta, \phi}\big(\mathbf{X}_{:,i}, \hat{\mathbf{X}}_{:,i}\big)>+\lambda \sum_{j}a_j^2\bigg),\\
    &=\argmin_{f, g}\bigg(<\Delta_{\theta, \phi}\big(\mathbf{X}_{:,i}, g(f(\mathbf{X}_{:,i}))\big)>+\lambda \sum_{j}a_j^2\bigg),\label{eq:reconstructionlos}
\end{align}
where $<\cdot>$ is the average operator and $\lambda$ is the hyper-parameter that weights the regularization term, $a_j$.  The higher the value of $\lambda$ the greater the capacity penalty. Note that $j$ runs across the hidden layers of the neural networks $f, g$. 
Finding the optimal neural networks $f$ and $g$ involves updating their respective learning parameters $\theta, \phi$ in each hidden layer of the neural networks so that the loss $\Delta_{\theta, \phi}$ is smaller than a given limit. In this work, the mean absolute  loss:
	\begin{equation}\label{eqn:3.1.1b}
		<\Delta_{\theta, \phi}\big(\mathbf{X}_{:,i}, \hat{\mathbf{X}}_{:,i}\big)> = \frac{1}{M} \sum_{i=1}^{M}\bigg|\mathbf{X}_{:,i}-\hat{\mathbf{X}}_{:,i}\bigg|
	\end{equation}
is used at the place of $<\cdot>$.

\subsection{Performance metrics}
Performance metrics measure the ability of the overall quality of the model. A single performance measure is not enough to validate an autoencoder, therefore, different measures are performed and evaluated.  The confusion matrix as shown in  Table \ref{confusion Table}  generates more meaningful measures to find the detection accuracy, precision, recall, and F1 score. 
True Normal, $TN$ represents the number of observations in the normal class that are predicted as normal by the model (i.e. below the threshold). True Anomaly, $TA$ is the number of observations in the anomaly class
that are predicted as an anomaly, and are above the threshold. False Normal, $FN$ is the number of anomalous observations  that are below the threshold (i.e. predicted as normal classes). False Anomaly,  $FA$ is the number of normal observations that are above the threshold (i.e. predicted as an anomaly).
\begin{table}
	\centering
	\begin{tabular}{|p{4cm}p{3.6cm}p{3.6cm}|} 
		\multicolumn{3}{c}{Classification Class Distribution } \\
		\hline 	\hline
		&Actual Normal & Actual Anomaly\\
		\hline
		Predicted Normal & \cellcolor{green!20} True Normal &\cellcolor{red!20}False Normal\\
		\hline
		Predicted Anomaly & \cellcolor{red!20}False Anomaly& \cellcolor{green!20}True Anomaly\\
		\hline
	\end{tabular}
\captionsetup{type=table} 
\caption{Two class classification confusion matrix representation }
\label{confusion Table}
\end{table}
The classification accuracy measures the general performance of the model by producing the ratio of true prediction (true normal and true anomaly) out of the total number of prediction:
	\begin{eqnarray}\label{accuracy}
		Accuracy = \frac{TN + TA}{ TA + TN +FA + FN}.
	\end{eqnarray}
The Precision is the ratio of the true anomaly divided by the total number of observations above the threshold (i.e. number or anomalies predicted):
	\begin{eqnarray}\label{Precision}
		Precision= \frac{TA}{TA +FA}.
	\end{eqnarray}  
The False positive rate, $FPR$ refers to the ratio of the normal samples above the threshold to the actual number of anomalous samples:
	\begin{eqnarray}\label{FPR}
		FPR  = \frac{FA}{FA +TN}.
	\end{eqnarray}
True positive rate, $TPR$ also known as sensitivity or recall, is the ratio of the number of anomalous samples above the threshold to the actual number of samples in the anomaly class:
	\begin{eqnarray}\label{TPR}
		TPR  = \frac{TA}{TA +FN}.
	\end{eqnarray}
Specificity is a measure obtained from the outcome of the confusion matrix which gives the ratio of true anomaly to the total negative class in the sample:  
	\begin{eqnarray}\label{Specificity}
		\text{Specificity} = \frac{TN}{TN + FA}.
	\end{eqnarray}
F-measure or $F_{1}$-score gives the harmonic mean between recall and  precision of the classifier. High F-measure indicates a better performance of the classifier with no false alarm:
	\begin{eqnarray}\label{FSCORE}
		F\text{-measure} = 2\left(\dfrac{\text{Precision}\times \text{Recall}}{\text{Precision} + \text{Recall}}\right).
	\end{eqnarray}

\section{Proposed label assisted autoencoder}
\label{chap:proposed}
\subsection{The architecture of the proposed model}
\label{arch}
\begin{figure}
	\includegraphics[width=1.2\textwidth]{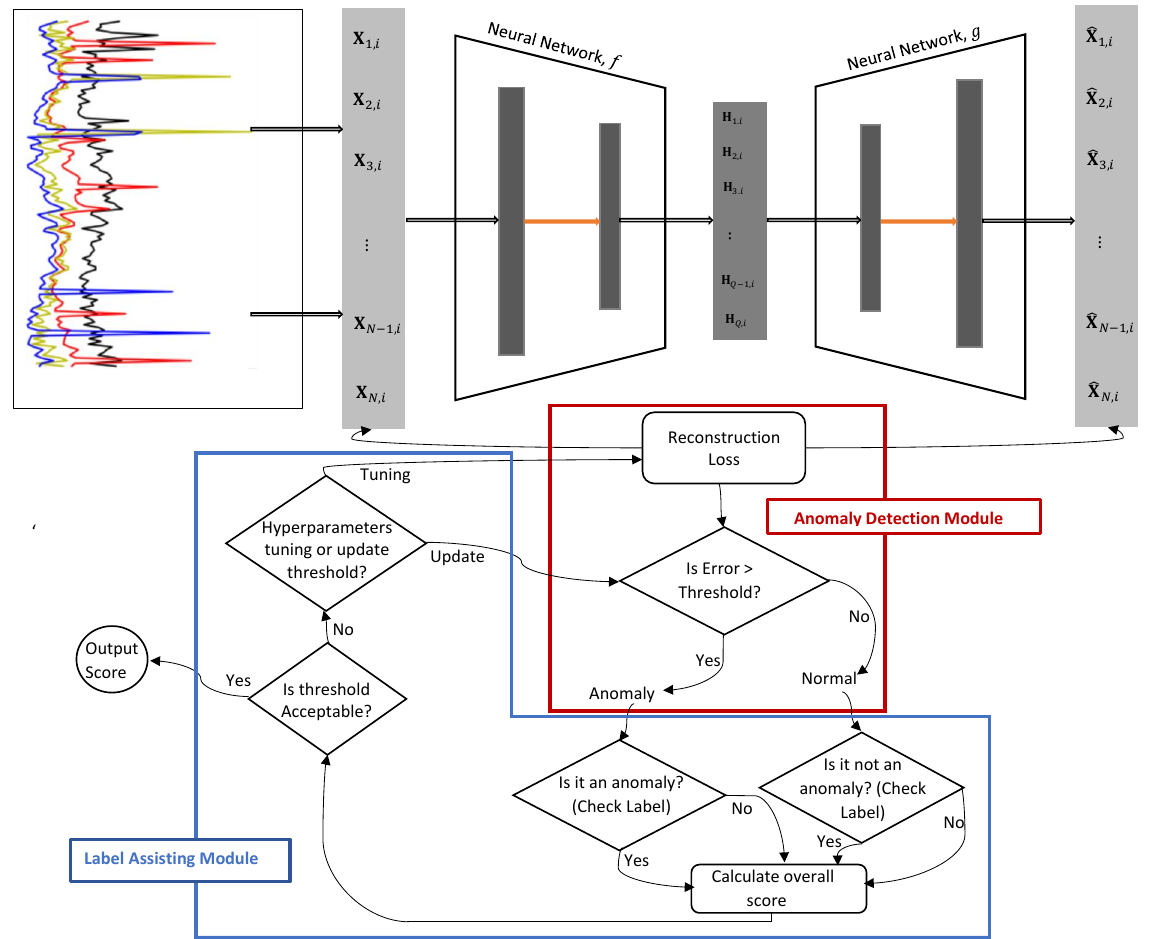}
	\caption{Overview of the proposed model. Each observation goes through the autoencoder for training, the reconstruction error is measured and then a threshold is used to make the decision whether the observation is an anomaly or not. The labelling assistance module then takes over to check if each of the observations is labelled, and then the label is used to check the veracity of the corresponding anomaly classification. A consensus is then reached whether training should stop or whether the threshold should be updated or the training should  continue with the search for hyper-parameters.}
	\label{fig:pmodel} 
\end{figure}

The encoder takes as input a  high-dimensional input data here a fixed-size vector and reaches the latent space through mapping it to a a low-dimensional representational vector. The decoder reconstructs the input data from the reduced representation in the latent space. The final reconstruction error is used to set a threshold to detect anomalies. An additional computation block is added to check from a set of labelled data if the threshold is acceptable to satisfy the required precision. This is a kind of validation block that uses the threshold to decide if the threshold should be changed or if the autoencoder should be trained to further minimize the reconstruction loss. The entire architecture of the proposed assisted autoencoder  is depicted in Figure \ref{fig:pmodel} and a detailed description is provided below.
\begin{enumerate}
    \item 
	The dataset is made up of observations and features which result in a 2-dimensional array of size $N \times M$, where $N$ and $M$ represent the number of features and  the number of observations respectively. All features of each observation are collected and received as input in the algorithm called nodes.
    \item  
	 The input data of size $N \times M$ is collected and reduced to a latent form of size $N \times Q$ where $Q < M$, while the $N \times Q$ set is fed into the decoder to produce a predicted output of size $N \times M$. Appendix \ref{Appendice} shows the parameters and the architecture of the deep neural networks trained in the encoder and decoder. The number of filters, size of filters and layers are displayed in the encoding and decoding phases.
     \item 
	After the decoding process, a reconstruction error is produced for each data point. The reconstruction error refers to the measure of how much the reconstructed input deviates from the original input.
	A threshold is set as a decision point to decide the acceptable amount of deviation, and the features of the  observation that go beyond this threshold is classified as an anomaly. The observation that is below this threshold is a normal data without anomaly.  
   \item The label assisting module then takes over to verify if each of the observations is labelled, then the label is used to verify the veracity of the corresponding anomaly classification. The labelled observations are then checked and an agreement is reached if the threshold is satisfied to obtain the desired precision to detect anomalies, if not the threshold is updated or the model is further trained to find the best hyper-parameters for the given threshold.
 \end{enumerate}
To illustrate the above steps, an example of a scenario is described as follows.  Assume  the input data is $\mathbf{X}_{:,i}$ with $N$ features and their corresponding predictions given as:
\begin{align}
  \mathbf{X}_{:,i}&=  \mathbf{X}_{1,i}, \mathbf{X}_{2,i}, \mathbf{X}_{3,i}, \cdots, \mathbf{X}_{N-1,i}, \mathbf{X}_{N,i}\\
                  &=(0.1, 0.2, 0.3, \cdots, 0.4, 0.5)\\
    \hat{\mathbf{X}}_{:,i}&=  \hat{\mathbf{X}}_{1,i}, \hat{\mathbf{X}}_{2,i}, \hat{\mathbf{X}}_{3,i}, \cdots, \hat{\mathbf{X}}_{N-1,i}, \hat{\mathbf{X}}_{N,i}\\
                  &=(0.6, 0.21, 0.32, \cdots, 0.61, 0.53)               
\end{align}
 The value of each of these features indicates the importance of the feature in the anomaly classification; e.g. the running time of a generator is more important than the generator capacity. The reconstruction loss  for each feature is then calculated as in Equation~\ref{eq:reconstructionlos}:
 \begin{align}
      L_{:,i}&=|\mathbf{X}_{:,i}-\hat{\mathbf{X}}_{:,i}|\\
             &= (0.5, 0.01, 0.02, 0.21, 0.03).
 \end{align}
 Assume the maximum reconstruction loss is set to $0.2$, then the absolute error classifies the sample under the normal category while if a priority is given to the important feature, say $\mathbf{X}_{N-1,i}$ then the sample would be labelled as an anomaly since the reconstruction loss of $\mathbf{X}_{N-1,i}$ is $L_{N-1, i}=0.21$ which is beyond $ 0.2$. Figure \ref{fig:recons} illustrates the reconstruction loss scenario adopted in this example.

\subsection{Training and testing phases}
Both the training  and testing phases are equally important when it comes to the generalisation of the model. The dataset is divided into training and testing, depending on the proportion of choice, in our case, we used the ratio of 3:1 for training and testing respectively. The training phase is divided into two folds. Firstly, minimising the reconstruction error is the focus of the training  for the reconstructed outputs to converge to the input samples.  Secondly, the reconstruction error is calculated for each data point to find an optimal threshold  for  anomaly detection.

\begin{figure}
	 	\centering
	 	\includegraphics[width=0.65\linewidth]{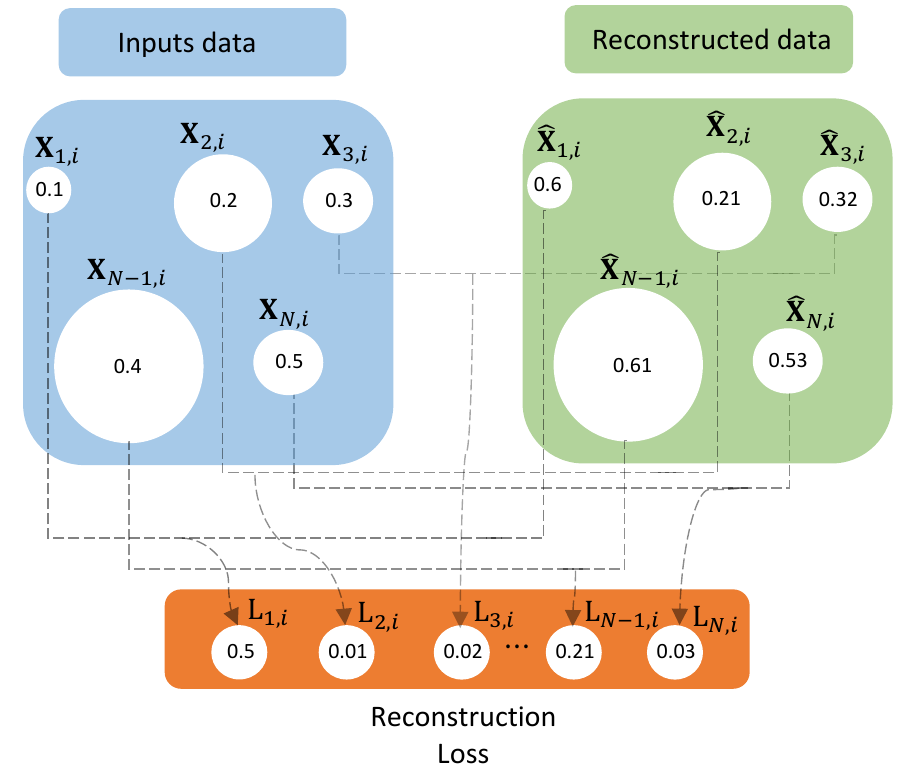}
	 	\caption{Computing reconstruction loss for $\mathbf{X}_{:,i}=(\mathbf{X}_{1,i}, \mathbf{X}_{2,i}, \mathbf{X}_{3,i}, \cdots, \mathbf{X}_{N-1,i}, \mathbf{X}_{N,i})$ using the predicted output $\hat{\mathbf{X}}_{:,i}=  (\hat{\mathbf{X}}_{1,i}, \hat{\mathbf{X}}_{2,i}, \hat{\mathbf{X}}_{3,i}, \cdots, \hat{\mathbf{X}}_{N-1,i}, \hat{\mathbf{X}}_{N,i})$.}
	 	\label{fig:recons}
	 \end{figure}

The training phase consists of normalizing the dataset so that all the features are reduced to a common scale without distorting the differences in the range of the values. In this work, the mathematical measure used to normalize the data is given by:
	\begin{equation}\label{scale}
		\mathbf{X}_{:,\mathrm{iscaled}} = \frac{\mathbf{X}_{:,i} - \mathbf{X}_{:,\mathrm{min}}}{\mathbf{X}_{:,\mathrm{max}}- \mathbf{X}_{:,\mathrm{min}}},
	\end{equation}
where $ \mathbf{X}_{:,\mathrm{min}}$ and $ \mathbf{X}_{:,\mathrm{max}}$ represent the data point  with  the minimum and maximum entries respectively. This is followed by dividing and separating the anomalous samples from the training set so that the algorithm learns to reconstruct the norms.  Once the training is completed, a reconstruction loss between the input and output is calculated and a backpropagation strategy is applied to adjust the weights and parameters of the model. 
The testing phase checks the performance of the model on the unseen test data using the threshold obtained from the training phase. The whole process is described in Figure~\ref{fig:recons}.
	 
\section{Datasets}
\label{chap:dataset}
\subsection{Data description}
\begin{figure}
	\centering
	\includegraphics[width=0.6\linewidth]{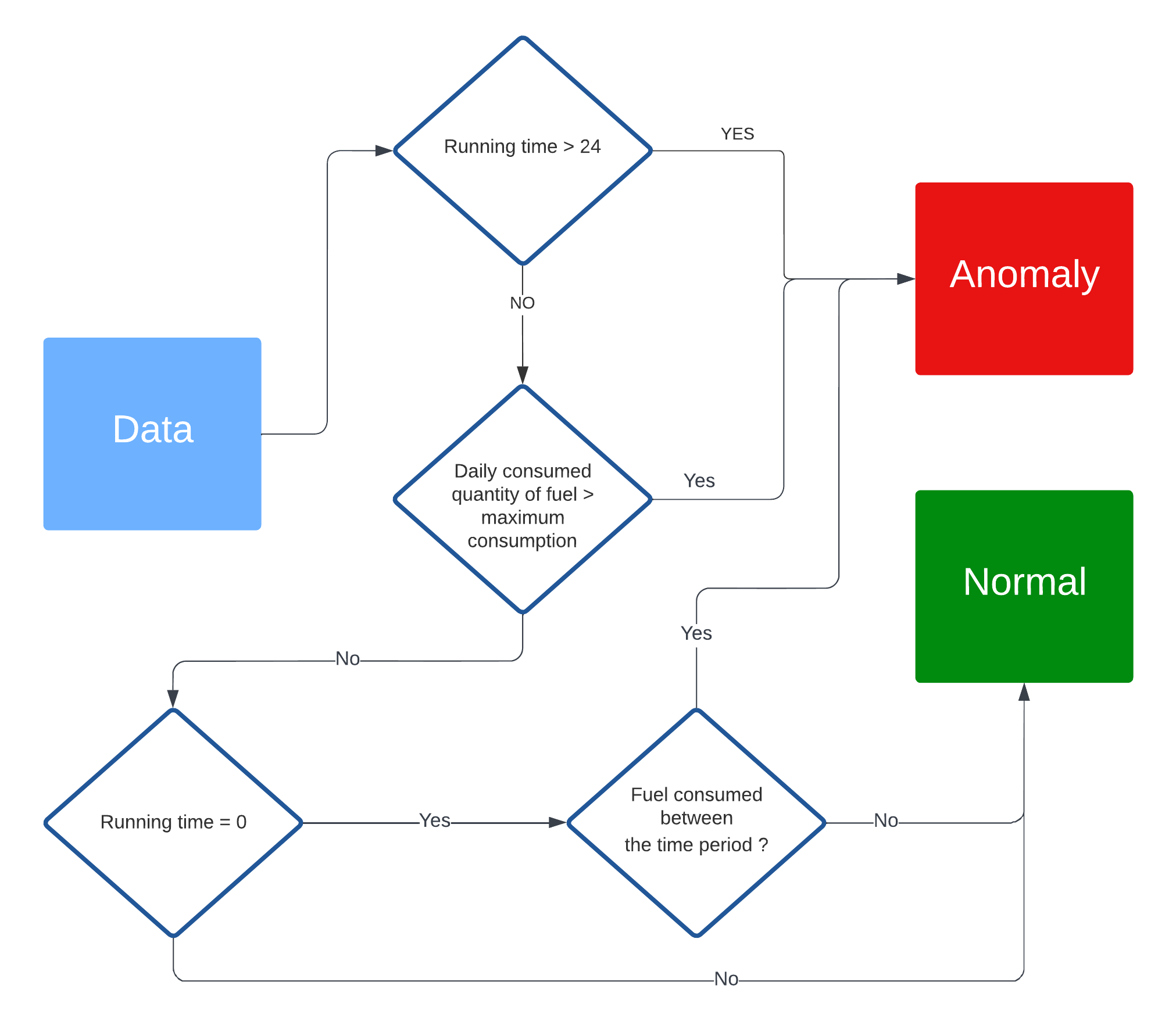}
	\caption{Flowchart showing how the labels are decided: For a data sample to receive the anomaly tag 1, it has to demonstrate at least one of the three anomaly indicators listed, otherwise it is given the normal tag 0.}
	\label{fig:classes}
\end{figure}

\begin{figure}
	\centering
	\includegraphics[width=0.7\linewidth]{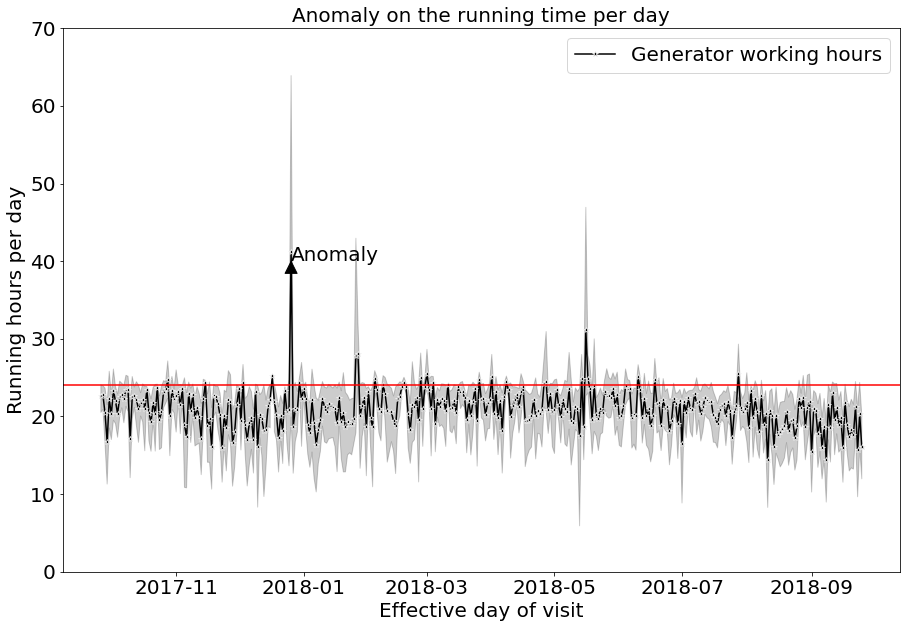}
	\caption{Observed anomaly in the number of working hours in a day for a single generator. For example, all data samples above 24 h threshold show  anomalies in the running time of the generator and these samples are assigned the label $1$, because it is known that one day only has 24 hours.}
	\label{fig:Anomaly}
\end{figure}

\begin{figure}
	 	\centering
	 	\includegraphics[width=0.9\linewidth]{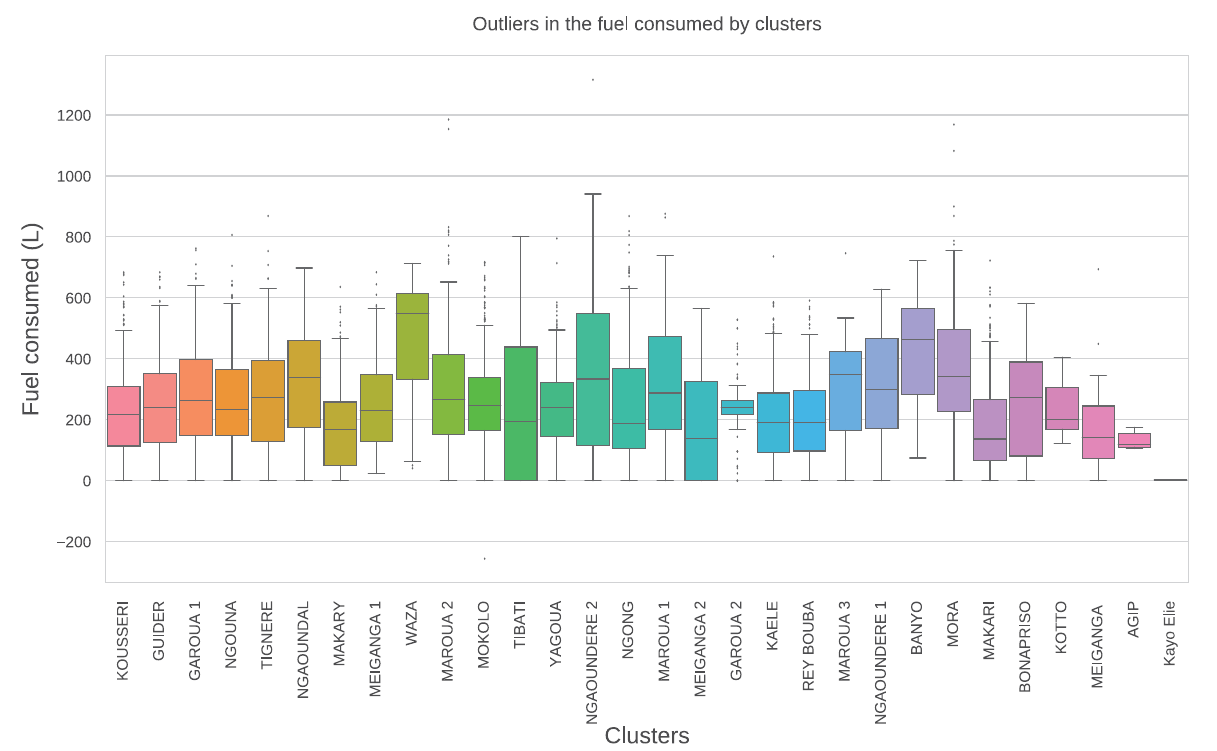}
	 	\caption{Fuel consumed per cluster showing the degree of anomalies in the dataset.}
	 	\label{fig:clusters}
	 \end{figure}

The dataset used in this paper is gathered from a Telecom base station management company in Cameroon named TeleInfra and was subject to previous studies for anomaly detection in \cite{mulongo2020anomaly}. The dataset is collected over the period of one year, i.e. from September 2017 to September 2018. It consists of $6010$ observations from various base stations in 46 towns and villages (known as clusters) across Cameroon.  These stations mainly rely on generators as the main supply of power. The dataset also consists of 17 variables which are defined in both numerical and categorical
forms. A detailed description of each variable is shown in Table \ref{tab: var}.
\setlength{\arrayrulewidth}{0.2mm}
\renewcommand{\arraystretch}{1.1}
\begin{table}
	\centering
	\begin{tabular}{|p{6cm}p{9cm}|}
		\hline
		\multicolumn{2}{|c|}{ Feature Description} \\
		\hline\hline
		CONSUMPTION HIS & The total fuel consumed between a specific period of time before the next refuelling is done.  \\
		\hline
		CONSUMPTION\_RATE & The number of litres the generator consumes
		per hour. \\
		\hline
		Cluster & The cities  where the generator sites are located\\
		\hline
		CURRENT HOUR METER GE1 & The hour meter reading of the generator. \\
		\hline
		Site Name & Name of the site where each generator is located \\
		\hline
		EFFECTIVE\_DATE\_OF\_VISIT & The date of meter reading, refuelling and recording   \\
		\hline
		PREVIOUS\_DATE\_OF\_VISIT & The previous date of visit  \\
		\hline
		Months &  The month when the reading was taken \\
		\hline
		NUMBER\_OF\_DAYS & The number of days before the next refuelling process.\\
		\hline
		GENERATOR\_1\_CAPACITY\_(KVA)
		&  The capacity of the generator \\
		\hline
		POWER TYPE & Type of power used in the power plant  \\
		\hline
		PREVIOUS HOUR METER G1 & The previous meter reading of the generator. \\
		\hline
		PREVIOUS\_FUEL\_QTE & The total quantity of fuel left inside the generator tank on the previous date of the visit. \\
		\hline
		QTE\_FUEL\_FOUND & The quantity of fuel found inside the generator tank before refuelling is done. \\
		\hline
		QTE\_FUEL\_ADDED & The quantity of fuel added to the generator
		during refuelling process. \\
		\hline
		TOTALE\_QTE\_LEFT & Quantity left in the generator after refuelling.\\
		\hline
		RUNNING\_TIME & The total number of hours the generator
		worked before the next refuelling is done \\
		\hline
	\end{tabular}
	\caption{Description of the different features in the dataset.}
	\label{tab: var}
\end{table}
Anomalies are observed in different features of the dataset and the  observed anomalies are classified based on three indicators: (1) for a given time period if the generator running time is zero and the quantity of fuel consumed is not zero. (2) when the running time per day is more than 24 hours, and (3) when the daily consumed quantity of fuel is more than the maximum consumption a generator can consume.
For a data sample to receive the anomaly tag 1, it has to demonstrate at least one of the three anomaly indicators listed, otherwise it is given the normal tag 0. A full workflow of the entire labelling process is illustrated in Figure \ref{fig:classes}. 
During the labelling process, output variables are assigned labels $0$ and $1$ representing the normal and anomaly classes respectively. For a single generator, Figure \ref{fig:Anomaly} illustrates the plot of the working hours per day, for example, all data samples above 24 h threshold show  anomalies in the running time of one of the generators and these samples are assigned the label $1$ because it is known that one day only has   24 hours. The 6010 observations are curated to remove missing samples leaving 5902 observations with all information. Out of $5905$ observations, it is observed that $3832$ samples are labelled as normal and $2073$ samples as abnormal, resulting in 64.8\% normal samples and 35.1\% abnormalities in the entire dataset. Figure \ref{fig:clusters}  shows all the clusters  and their respective total fuel consumption including the degree of  anomalies in the entire dataset.

\subsection{Feature importance} \label{Feature importance}
\begin{figure}
	\centering
	\includegraphics[width=0.7\linewidth]{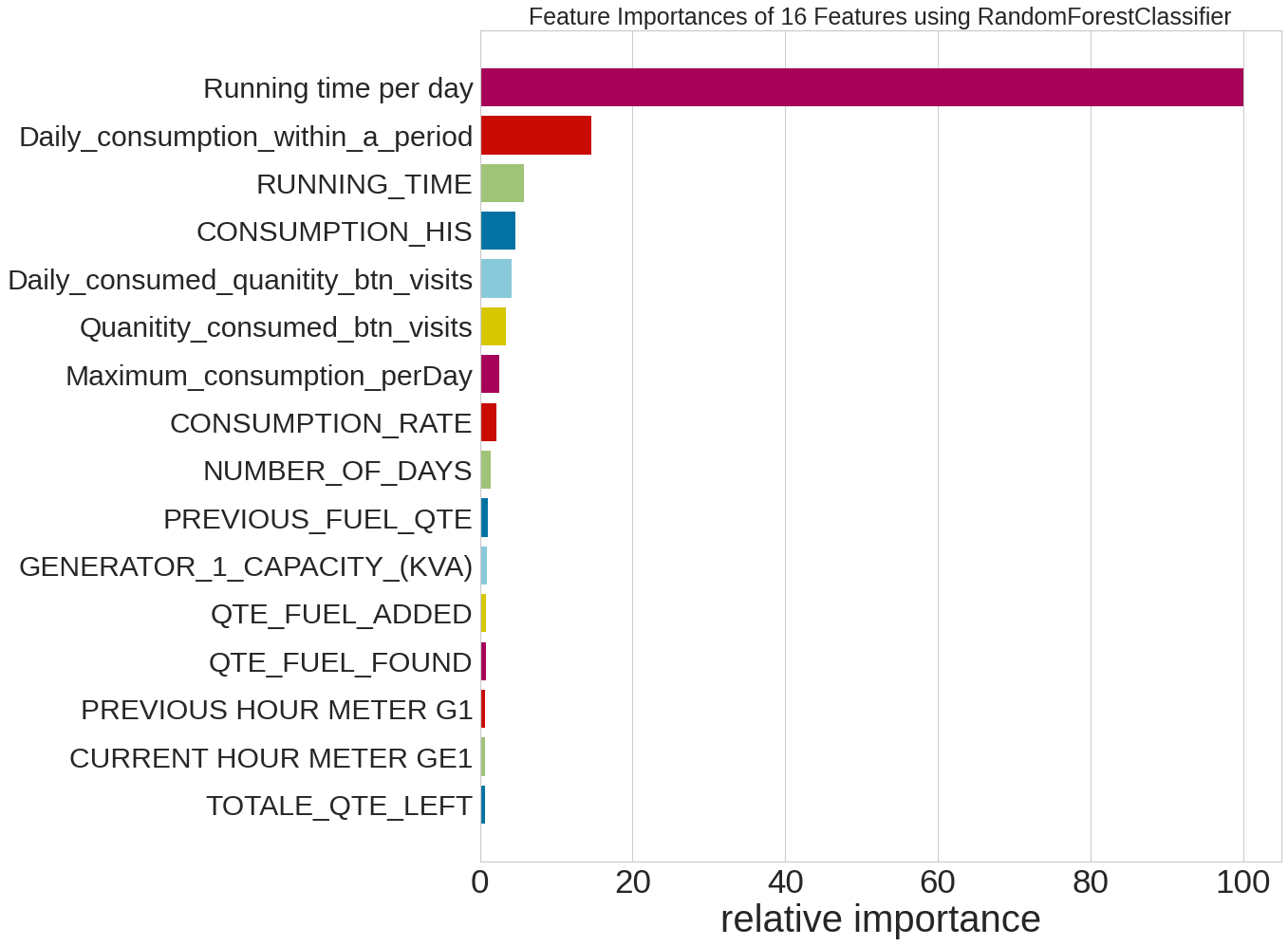}
	\caption{Feature importance for the 16 variables fitted using Random Forest Classifier.  The feature \enquote{Running time per day} has the greatest influence on the output and can be coined as the most important feature in the dataset.}
	\label{fig:featureimportance}
\end{figure}

Feature selection is performed by fitting the data using a random forest classifier with 16 features. Note that one could use any other method to find the most important feature. Since the relative importance of the most important feature is too high (100\% as shown in Figure \ref{fig:featureimportance}) compared to other features, any algorithm will predict the same feature as the most important. Figure \ref{fig:featureimportance} shows that the feature \enquote{Running time per day} has the greatest influence on the output and can be coined as the most important feature in the dataset. Even though it is followed by the
\enquote{Daily consumption within a period} the huge difference between the \enquote{Running time per day} and the remaining features shows that 
 priority should be given to the feature \enquote{Running time per day}  when considering its reconstruction error for anomaly detection.

\subsection{Correlation}
The correlation matrix is used to visualize the linear relationship between two variables. The values produced by the covariance matrix range from -1 to 1, where 1 indicates a strong positive linear relationship, -1 indicates a strong negative linear relationship and 0 indicates that there is no linear relationship between the variables.
Figure \ref{fig:corr} shows that the key variable \enquote{Running time per day} and the variable \enquote{Daily consumption within a period} has a strong positive correlation, which is reasonable since the daily  quantity of fuel consumed by a generator is  a function of  the running time. A strong correlation is also observed between the three pairs of features; \enquote{Total quantity of fuel after refilling}, \enquote{Quantity of fuel found} and the \enquote{Previous quantity of fuel recorded}. However, the feature \enquote{Previous hour meter G1} has no significant correlation with any of the features.

\section{Results}
\label{chap:results}
The results are observed from a number of different evaluation metrics.
\subsection{Training}
\begin{figure}
		\centering
		\includegraphics[width=0.55\linewidth]{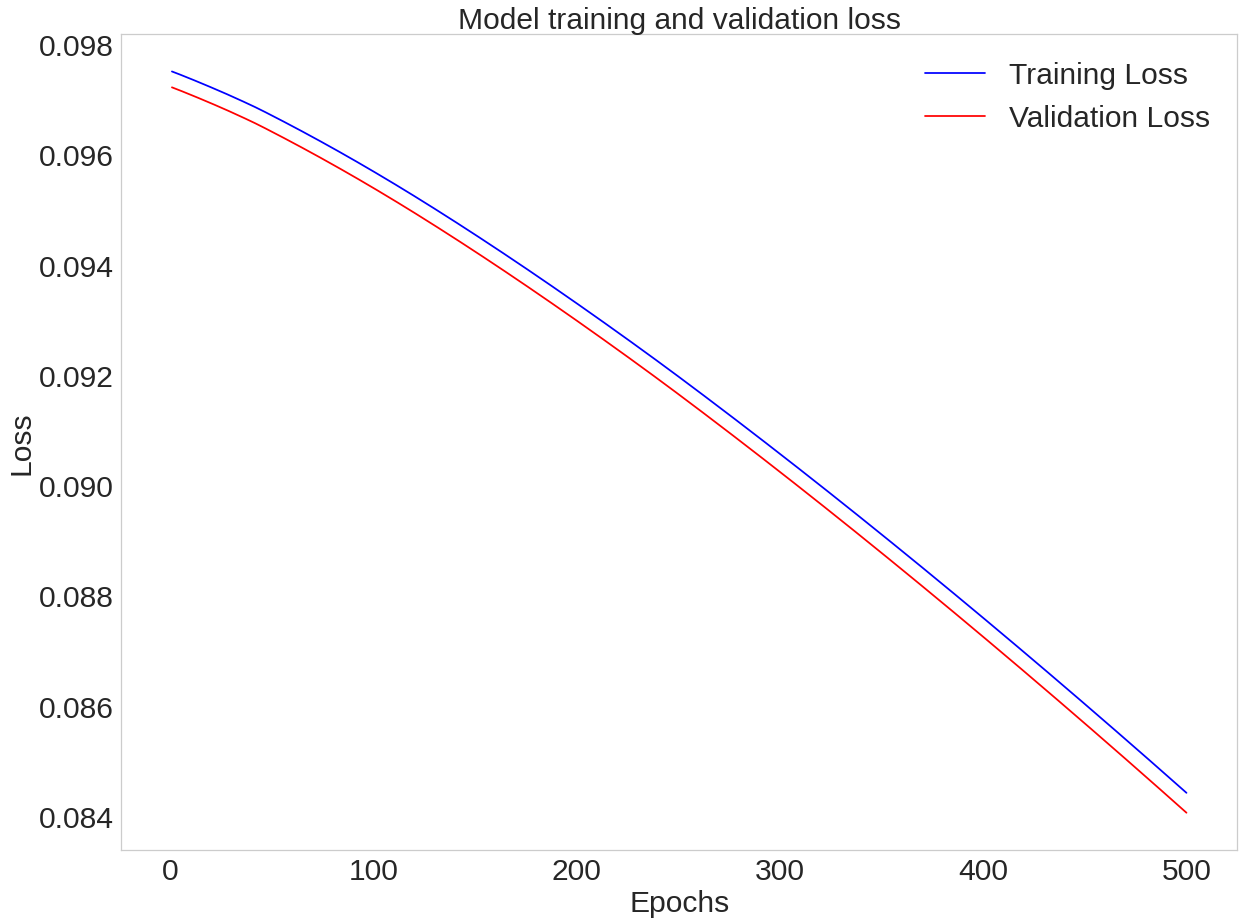}
		\caption{Training and validation loss as a function of the number of iterations.  The mean absolute error is used to measure the loss.}
		\label{fig:loss}
	\end{figure}

\begin{figure}
	\begin{subfigure}[b]{0.5\textwidth}
		\centering
		\includegraphics[width=0.8\linewidth]{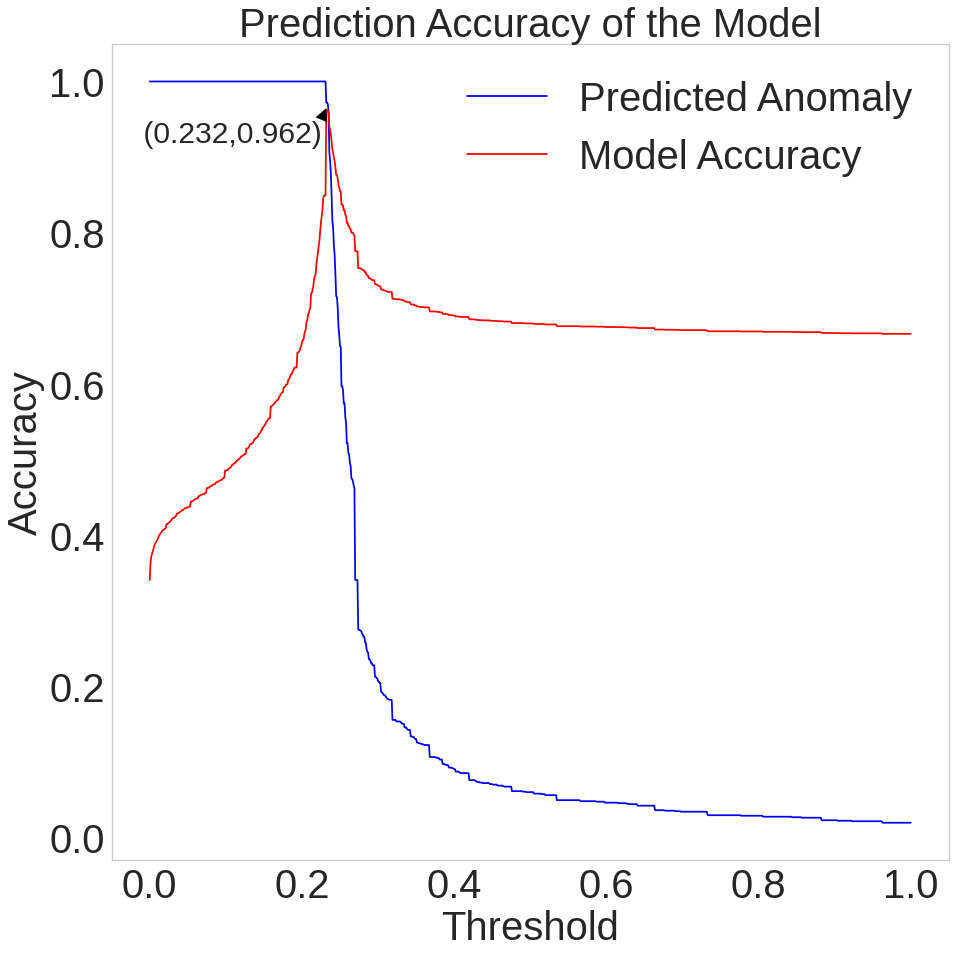}
		\caption{Threshold detection}
		\label{Threshold}
	\end{subfigure}
	\hfill
	\begin{subfigure}[b]{0.5\textwidth}
		\centering
		\includegraphics[width=0.8\linewidth]{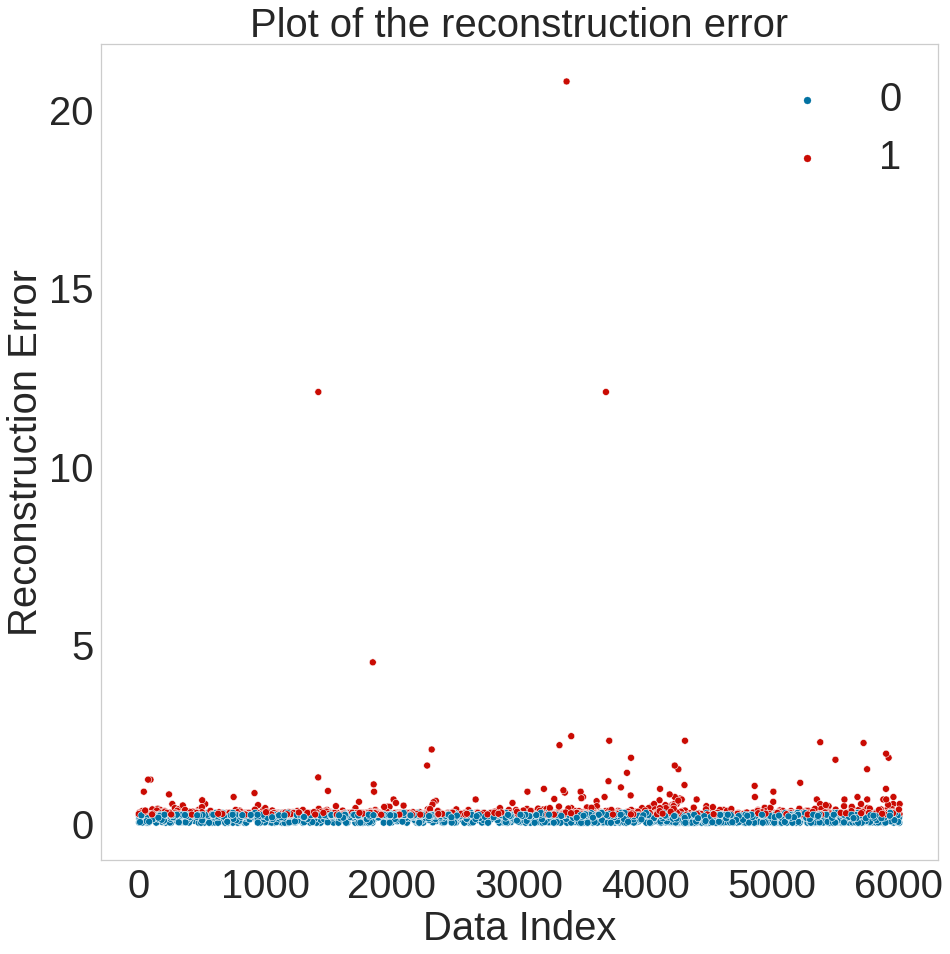}
		\caption{Reconstruction Error}
		\label{Recon}
	\end{subfigure}
	\caption{Threshold detection  (\ref{Threshold}) using the reconstruction error (\ref{Recon}). The threshold is  used to categorise  anomalies from mild to extreme using their reconstruction error.}
	\label{Thresh}
\end{figure}

	In order to assess the performance of our model during
	training, we set aside 10\% of the training data as the validation set. The training and validation loss are seen in Figure \ref{fig:loss} as it trends down different epochs. We observe that the validation loss is below the training loss. However, this can not influence the predicted accuracy with further hyper-parameter tuning since the difference between the two errors is negligible. 
The training loss accesses the error in the model during training.
Finding the model's appropriate threshold  requires testing the model with the entire test dataset. The confusion matrix is used to get the model accuracy and True positive rate (TPR). They are plotted over a range of thresholds as seen in Figure \ref{Threshold} and the level at which they both attain an average maximum point is the best threshold for the model. As seen in  Figure \ref{Threshold}, the best threshold is 0.232 with an accuracy score of 0.962. However, if the sensitivity (TPR) or predicted anomaly is a priority for the organization without minding the cost they may secure while sorting out the False normal (predicted anomalies that are actually), the threshold could then be reduced to 0.231. This is the point where all anomalies are predicted (i.e. predicted anomalies are at 100\%) with a higher false normal (FN) which  has an effect on  the overall model accuracy. 

The threshold is also used to categorise these anomalies from mild to extreme using their reconstruction error. Figure \ref{Recon} shows the reconstruction error. The samples with extreme threshold are prioritised when one seeks to find the reasons for the presence of anomalies. 

\subsection{Model performance} 
\begin{figure}
	\centering
	\begin{subfigure}{0.45\linewidth}
	\centering
	\includegraphics[width=0.8\linewidth]{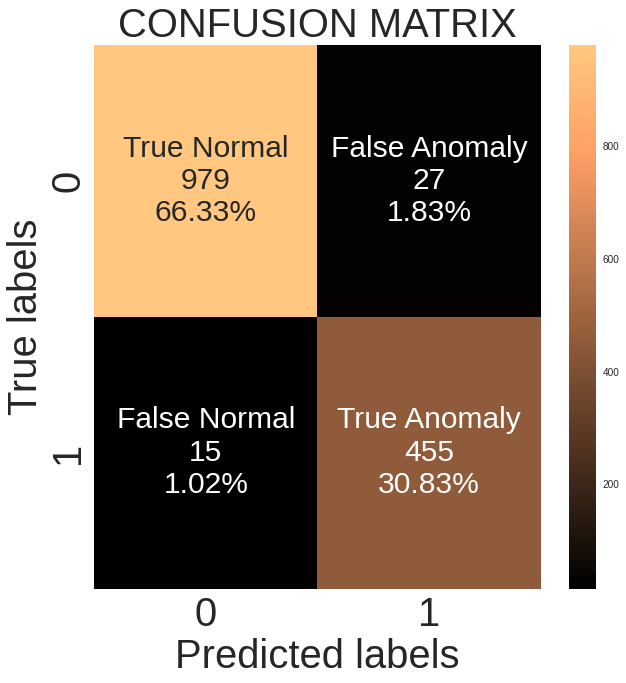}
	\caption{Threshold = 0.232}
	\label{fig:cm}
	\end{subfigure}
	\begin{subfigure}{0.45\linewidth}
	\centering
	\includegraphics[width=0.8\linewidth]{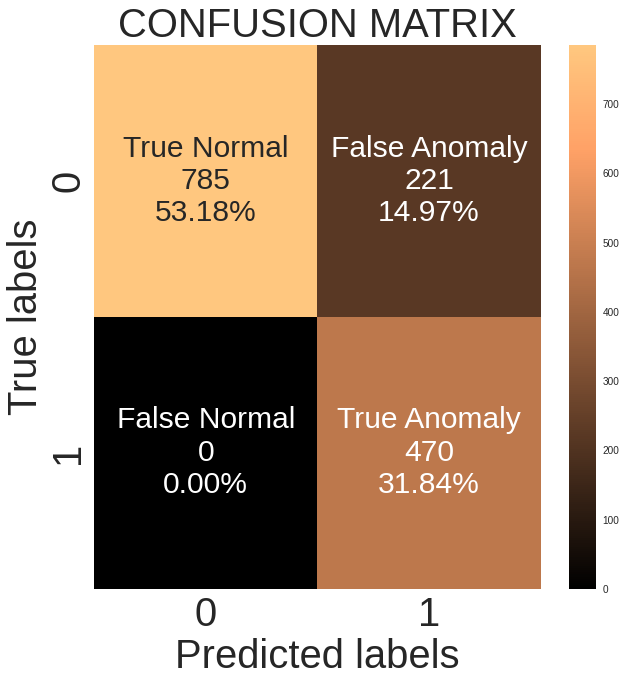}
	\caption{Threshold = 0.231}
	\label{fig:cm1}
	\end{subfigure}
	\caption{Detection of results based on confusion matrix.}
	\label{fig:Cm}
\end{figure}

The total number of test samples is 1,476 with 1,006 normal samples and the remaining 470 are  anomalous samples. Figure \ref{fig:Cm} illustrates the performance of the proposed  model at different thresholds based on the confusion matrix. 
The proposed model is able to detect a total of 455
anomalous samples correctly out of the 470 samples with the anomaly label, this accounts for 96.8\% (TPR) of the total anomaly samples. The model also detected a
total number of 979 normal samples correctly out of
1006.  The 
model incorrectly classified 15 normal
samples as  anomalies (FN) and 27 anomalies as normal samples (FA). These results show an accuracy of 97.15\%, a precision of 94.40\%, a recall of 96.81\%, a specificity of 97.31\% and a F1-score of 95.59\%. Table \ref{Performance} provides a summary of the performance metrics.
\subsection{Comparison with other models using the same datatset}
Table \ref{comparison} shows the performance of the proposed model compared to that discussed in \cite{mulongo2020anomaly}. We used the latter paper to compare  our work with because both are implemented  using the same Teneifera dataset for anomaly detection. The proposed label assisted autoencoder has the best performance with an accuracy of 97.2\% and a recall of 96.8\%. The
multi-layer perception proposed in \cite{mulongo2020anomaly} shows the most competitive performance  with a higher F1-score, specificity and precision. However, the label assisted autoencoder  is  flexible since if we adjust the  threshold,  the recall will increase at the cost of specificity and overall accuracy.

\subsection{Anomaly classification} 
The reconstruction error for each data sample differs from one another (as seen in Figure \ref{Recon}) and provides an opportunity to classify these predicted anomalies according to their reconstruction error. In this work, 4 classes $A, B, C$, and $D$ are considered. Class $A$ represents anomalies that are slightly above the threshold, class $B$ represents  anomalies that are above twice the threshold, class $C$ represents  anomalies that are above four times the threshold, and class D represents  anomalies that are above eight times the threshold. This implies that each class has twice a threshold  compared to its predecessor. Table \ref{anomaly} shows the classes and   their corresponding thresholds showing that 28.25\% of the test dataset belongs to the anomaly category of class $A$, 2.03\% belongs to class $B$, 0.20\% belongs to class $C$ and 0.34\% belongs to $D$.

\subsection{Discussion}
\begin{figure}
	\centering
	\includegraphics[width=0.65\linewidth]{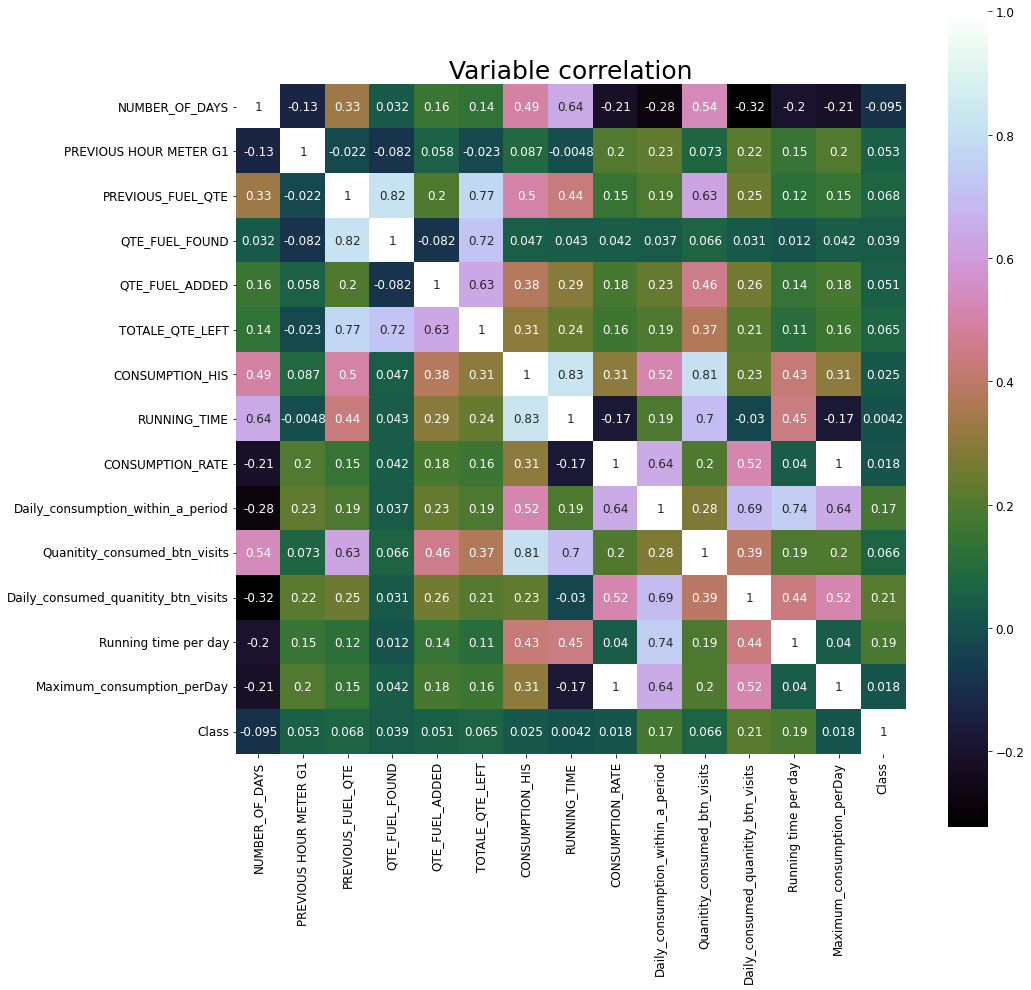}
	\caption{Correlation matrix of all numerical features. The \enquote{Running time per day} and \enquote{Daily consumption within a period} have a strong positive correlation, which is reasonable since the daily  quantity of fuel consumed by a generator is  a function of  the running time.}
	\label{fig:corr}
\end{figure}

Feature importance has played an important role in this model, the feature \enquote{Running time per day} has the most significant importance of 100, and \enquote{Daily consumption within a period} coming in the second place with an important measure of 16, from this we gave priority to the reconstruction error of the \enquote{Running time per day}. To further justify our choice, the correlation matrix in Figure \ref{fig:corr} shows a strong positive correlation of 0.74 between the two features.
Using the reconstruction error of the key variable, we were able to train and compare the proposed model with the work in \cite{mulongo2020anomaly}. A recall score of 96.8\% outperformed all the models proposed in \cite{mulongo2020anomaly} in detecting anomalies. Also, as shown in Table \ref{Performance}, our model shows a recall score of approximately 100\% with a decrease in the  overall accuracy of 85\% when the threshold is decreased to 0.231.
\section{Conclusion}
\label{chap:conclud}
Telecommunication industry is one of the dominant information communication technology industries that rely on a huge amount of electric power supply for their operations, and thus it is indispensable in their daily dealings. However, its availability in underdeveloped countries, particularly in Africa, has been a constant source of contention. Despite the industry's rise through the creation of base stations, they have had to turn to alternative energy sources such as the use of gasoline or diesel with generators, and the use of solar power, to name a few. 

TeleInfra telecommunication company established in Cameroon is one of such companies hooked on these challenges due to the state of power supply in the country. The telecommunication equipment that is fixed in different parts of the rural and urban areas in Cameroon requires an uninterrupted supply of electricity to achieve the goal of establishing strong and seamless communication channels in the country, however, the country's electrical generation is mostly based on hydropower (73\%), with perpetual power interruptions, particularly during the dry seasons when water levels are low \citep{erasmus}.
The consequence of the diversification to alternative sources of power, particularly the usage of generators posed another challenge of irregularities or anomalies in fuel consumption at the base stations due to the observed high consumption rate in the power generation plants. 
TeleInfra telecommunication company is faced with the challenge of unaccounted high fuel consumption for their operations at the base stations. Since they solely depend on generating plants as their major source of power supply, they necessarily have to continually refill these generators and these are done manually. Such activities are known to have emanated in possible cases of pilferage of fuel due to the observed anomalies in  fuel consumption. As a result, it is essential to investigate the likely factors contributing to the anomalies by collecting data on fuel consumption at each of the base stations for the purpose of minimizing the costs of operation. 

We have proposed a label assisted autoencoder-based deep-learning
technique for detecting anomalies in the fuel consumption datasets of the base station management company namely  TeleInfra. In our proposed model, an autoencoder is used to generate an encoded representation of the input features and construct the output from the encoded representation to look like the input features of the series of decoders. The maximum reconstruction error from the trained model is obtained from the training set and it is set as a threshold for detecting anomalies on the test dataset. The anomaly detector identifies each data sample from the testing set as anomaly when they exceed the threshold assigned. 
Results showed that our proposed model is highly efficient for reading anomalies with a detection
accuracy of 97.20\% and outperforms existing supervised learning models. The proposed model is flexible and the threshold is adjustable according to the need of the user which can classify anomalies from severe to mild.

This work opens future research possibilities which could involve using different variations of autoencoders such as long short term memory autoencoders and memory-augmented autoencoders combined with our proposed label assisted unit. The latter does not require the feature importance analysis for selecting the best reconstruction error.
\setlength{\arrayrulewidth}{0.2mm}
\renewcommand{\arraystretch}{1.1}
\begin{table}
	\centering
	\begin{tabular}{ |p{2cm}|p{2cm}|p{2cm}|p{2cm}| p{2cm}| p{2cm}|} 
		\hline
		\multicolumn{6}{c}{\cellcolor{gray!50}Model performance } \\
		\hline
		Threshold&Accuracy &F1-Measure & Recall & Precision & Specificity\\
		\cline{1-6}
		0.232&0.972&0.956&0.968&0.944&0.973\\
		\hline
		0.231&0.850&0.810&1.00&0.680&0.780\\
		\hline
	\end{tabular}
	\captionsetup{type=table} 
	\caption{Evaluation performance.}
	\label{Performance}
\end{table}
\setlength{\arrayrulewidth}{0.2mm}
\renewcommand{\arraystretch}{1.1}
\begin{table}
	\centering
	\begin{tabular}{p{3cm}|p{1.7cm}|p{1.5cm}|p{1.7cm}|p{1.2cm}|p{1.3cm}| p{1.7cm}|} 
		\hline
		\multicolumn{7}{c}{\cellcolor{gray!50}Model performance } \\
		\hline
		Paper&Techniques&Accuracy &F1-Score & Recall & Precision & Specificity\\
		\cline{1-7}
	Mulongo et al. \cite{mulongo2020anomaly}	&LR&0.708&0.811&0.709&0.943&0.699\\
		\cline{2-7}
		&SVM&0.949&0.962&0.962&0962&0.925\\
		\cline{2-7}
		&KNN&0.851&0.888&0.887&0.890&0.783\\
		\cline{2-7}
		&MLP&0.961&0.971&0.954&0.988&0.976\\
		\cline{1-7}
	Our Model&AE&0.972&0.956&0.968&0.944&0.973\\
		\hline
	\end{tabular}
	\captionsetup{type=table} 
	\caption{Comparison to similar models using the same dataset.}
	\label{comparison}
\end{table}
\begin{table}
	\centering
	\begin{tabular}{ |p{2cm}|p{2cm}|p{5cm}|p{4cm}|} 
		\hline
		\multicolumn{4}{c}{\cellcolor{gray!50}Categorizing Anomalies} \\
		\hline
		Class&Threshold &Predicted Number of Samples & Percentage of Test Data\\
		\cline{1-4}
		\cellcolor{red!10} A&0.232&417&28.25\%\\
		\hline
		\cellcolor{red!30}B&0.464&30&2.03\%\\
		\hline
		\cellcolor{red!50} C&0.928&3&0.20\%\\
		\hline
		\cellcolor{red!70} D&1.856&5&0.34\%\\
		\hline
	\end{tabular}
	\captionsetup{type=table} 
	\caption{Four categories of anomalies.}
	\label{anomaly}
\end{table}
\section*{Disclosure Statement}
We hereby state that no known competing financial interests or personal ties could have
influenced the research presented in this study.
\bibliography{bibliography}

\begin{thebibliography}{10}
\expandafter\ifx\csname url\endcsname\relax
  \def\url#1{\texttt{#1}}\fi
\expandafter\ifx\csname urlprefix\endcsname\relax\def\urlprefix{URL }\fi
\expandafter\ifx\csname href\endcsname\relax
  \def\href#1#2{#2} \def\path#1{#1}\fi

\bibitem{humar2011rethinking}
I.~Humar, X.~Ge, L.~Xiang, M.~Jo, M.~Chen, J.~Zhang, Rethinking energy
  efficiency models of cellular networks with embodied energy, IEEE network
  25~(2) (2011) 40--49.

\bibitem{lorincz2013renewable}
J.~Lorincz, I.~Bule, Renewable energy sources for power supply of base station
  sites, International Journal of Business Data Communications and Networking
  (IJBDCN) 9~(3) (2013) 53--74.

\bibitem{ayang2016power}
A.~Ayang, P.-S. Ngohe-Ekam, B.~Videme, J.~Temga, Power consumption: base
  stations of telecommunication in sahel zone of cameroon: typology based on
  the power consumption model and energy savings, Journal of Energy 2016.

\bibitem{espadafor2009analysis}
F.~J. Espadafor, J.~B. Villanueva, M.~T. Garc{\'\i}a, Analysis of a diesel
  generator crankshaft failure, Engineering Failure Analysis 16~(7) (2009)
  2333--2341.

\bibitem{mulongo2020anomaly}
J.~Mulongo, M.~Atemkeng, T.~Ansah-Narh, R.~Rockefeller, G.~M. Nguegnang, M.~A.
  Garuti, Anomaly detection in power generation plants using machine learning
  and neural networks, Applied Artificial Intelligence 34~(1) (2020) 64--79.

\bibitem{jinggang2009feasibility}
W.~Jinggang, K.~Ligai, D.~Meixia, Z.~Jin, G.~Xiaoxia, Feasibility analysis
  using natural source cooling the idc plant, in: 2009 Chinese Control and
  Decision Conference, IEEE, 2009, pp. 2579--2584.

\bibitem{goodfellow2016deep}
I.~Goodfellow, Y.~Bengio, A.~Courville, Deep learning, MIT press, 2016.

\bibitem{aggarwal2017introduction}
C.~C. Aggarwal, An introduction to outlier analysis, in: Outlier analysis,
  Springer, 2017, pp. 1--34.

\bibitem{demestichas2021comparative}
K.~Demestichas, T.~Alexakis, N.~Peppes, E.~Adamopoulou, Comparative analysis of
  machine learning-based approaches for anomaly detection in vehicular data,
  Vehicles 3~(2) (2021) 171--186.

\bibitem{hayes2015contextual}
M.~A. Hayes, M.~A. Capretz, Contextual anomaly detection framework for big
  sensor data, Journal of Big Data 2~(1) (2015) 1--22.

\bibitem{fahim2019anomaly}
M.~Fahim, A.~Sillitti, Anomaly detection, analysis and prediction techniques in
  iot environment: A systematic literature review, IEEE Access 7 (2019)
  81664--81681.

\bibitem{trinh2019detecting}
H.~D. Trinh, E.~Zeydan, L.~Giupponi, P.~Dini, Detecting mobile traffic
  anomalies through physical control channel fingerprinting: A deep
  semi-supervised approach, IEEE Access 7 (2019) 152187--152201.

\bibitem{omar2013machine}
S.~Omar, A.~Ngadi, H.~H. Jebur, Machine learning techniques for anomaly
  detection: an overview, International Journal of Computer Applications
  79~(2).

\bibitem{vishwanathan2002ssvm}
S.~Vishwanathan, M.~N. Murty, Ssvm: a simple svm algorithm, in: Proceedings of
  the 2002 International Joint Conference on Neural Networks. IJCNN'02 (Cat.
  No. 02CH37290), Vol.~3, IEEE, 2002, pp. 2393--2398.

\bibitem{peterson2009k}
L.~E. Peterson, K-nearest neighbor, Scholarpedia 4~(2) (2009) 1883.

\bibitem{nick2007logistic}
T.~G. Nick, K.~M. Campbell, Logistic regression, Topics in biostatistics (2007)
  273--301.

\bibitem{ruck1990feature}
D.~W. Ruck, S.~K. Rogers, M.~Kabrisky, Feature selection using a multilayer
  perceptron, Journal of Neural Network Computing 2~(2) (1990) 40--48.

\bibitem{bhattacharyya2011data}
S.~Bhattacharyya, S.~Jha, K.~Tharakunnel, J.~C. Westland, Data mining for
  credit card fraud: A comparative study, Decision Support Systems 50 (2011)
  602--613.

\bibitem{hasan2019attack}
M.~Hasan, M.~M. Islam, M.~I.~I. Zarif, M.~Hashem, Attack and anomaly detection
  in iot sensors in iot sites using machine learning approaches, Internet of
  Things 7 (2019) 100059.

\bibitem{said2020network}
M.~Said~Elsayed, N.-A. Le-Khac, S.~Dev, A.~D. Jurcut, Network anomaly detection
  using lstm based autoencoder, in: Proceedings of the 16th ACM Symposium on
  QoS and Security for Wireless and Mobile Networks, 2020, pp. 37--45.

\bibitem{hawkins2002outlier}
S.~Hawkins, H.~He, G.~Williams, R.~Baxter, Outlier detection using replicator
  neural networks, in: International Conference on Data Warehousing and
  Knowledge Discovery, Springer, 2002, pp. 170--180.

\bibitem{zhang2021power}
J.~Zhang, H.~Zhang, S.~Ding, X.~Zhang, Power consumption predicting and anomaly
  detection based on transformer and k-means, Frontiers in Energy Research
  (2021) 681.

\bibitem{munz2007traffic}
G.~M{\"u}nz, S.~Li, G.~Carle, Traffic anomaly detection using k-means
  clustering, in: GI/ITG Workshop MMBnet, Vol.~7, 2007, p.~9.

\bibitem{zhou2017anomaly}
C.~Zhou, R.~C. Paffenroth, Anomaly detection with robust deep autoencoders, in:
  Proceedings of the 23rd ACM SIGKDD international conference on knowledge
  discovery and data mining, 2017, pp. 665--674.

\bibitem{mishra2018generative}
A.~Mishra, S.~Krishna~Reddy, A.~Mittal, H.~A. Murthy, A generative model for
  zero shot learning using conditional variational autoencoders, in:
  Proceedings of the IEEE conference on computer vision and pattern recognition
  workshops, 2018, pp. 2188--2196.

\bibitem{johnson2017generative}
G.~R. Johnson, R.~M. Donovan-Maiye, M.~M. Maleckar, Generative modeling with
  conditional autoencoders: Building an integrated cell, arXiv preprint
  arXiv:1705.00092.

\bibitem{ge2019dual}
P.~Ge, C.-X. Ren, D.-Q. Dai, J.~Feng, S.~Yan, Dual adversarial autoencoders for
  clustering, IEEE transactions on neural networks and learning systems 31~(4)
  (2019) 1417--1424.

\bibitem{guo2017deep}
X.~Guo, X.~Liu, E.~Zhu, J.~Yin, Deep clustering with convolutional
  autoencoders, in: International conference on neural information processing,
  Springer, 2017, pp. 373--382.

\bibitem{luo2017convolutional}
W.~Luo, J.~Li, J.~Yang, W.~Xu, J.~Zhang, Convolutional sparse autoencoders for
  image classification, IEEE transactions on neural networks and learning
  systems 29~(7) (2017) 3289--3294.

\bibitem{ferreira2020recommendation}
D.~Ferreira, S.~Silva, A.~Abelha, J.~Machado, Recommendation system using
  autoencoders, Applied Sciences 10~(16) (2020) 5510.

\bibitem{sakurada2014anomaly}
M.~Sakurada, T.~Yairi, Anomaly detection using autoencoders with nonlinear
  dimensionality reduction, in: Proceedings of the MLSDA 2014 2nd workshop on
  machine learning for sensory data analysis, 2014, pp. 4--11.

\bibitem{wang2014generalized}
W.~Wang, Y.~Huang, Y.~Wang, L.~Wang, Generalized autoencoder: A neural network
  framework for dimensionality reduction, in: Proceedings of the IEEE
  conference on computer vision and pattern recognition workshops, 2014, pp.
  490--497.

\bibitem{michelucci2022introduction}
U.~Michelucci, An introduction to autoencoders, arXiv preprint
  arXiv:2201.03898.

\bibitem{wei2023lstm}
Y.~Wei, J.~Jang-Jaccard, W.~Xu, F.~Sabrina, S.~Camtepe, M.~Boulic,
  Lstm-autoencoder based anomaly detection for indoor air quality time series
  data, IEEE Sensors Journal.

\bibitem{bank2020autoencoders}
D.~Bank, N.~Koenigstein, R.~Giryes, Autoencoders, arXiv preprint
  arXiv:2003.05991.

\bibitem{erasmus}
E.~Muh, S.~Amara, F.~Tabet, Sustainable energy policies in cameroon: A holistic
  overview, Renewable and Sustainable Energy Reviews 82 (2018) 3420--3429.
\newblock \href {http://dx.doi.org/https://doi.org/10.1016/j.rser.2017.10.049}
  {\path{doi:https://doi.org/10.1016/j.rser.2017.10.049}}.

\end{thebibliography}
\newpage
\section*{Appendix A}
\label{Appendice}
Architecture of deep neural networks trained in the encoder and decoder. The number of filters, size of filters and layers are displayed in the encoding and decoding phases.
\begin{figure}[h]
	\centering
	\includegraphics[width=1.1\textwidth]{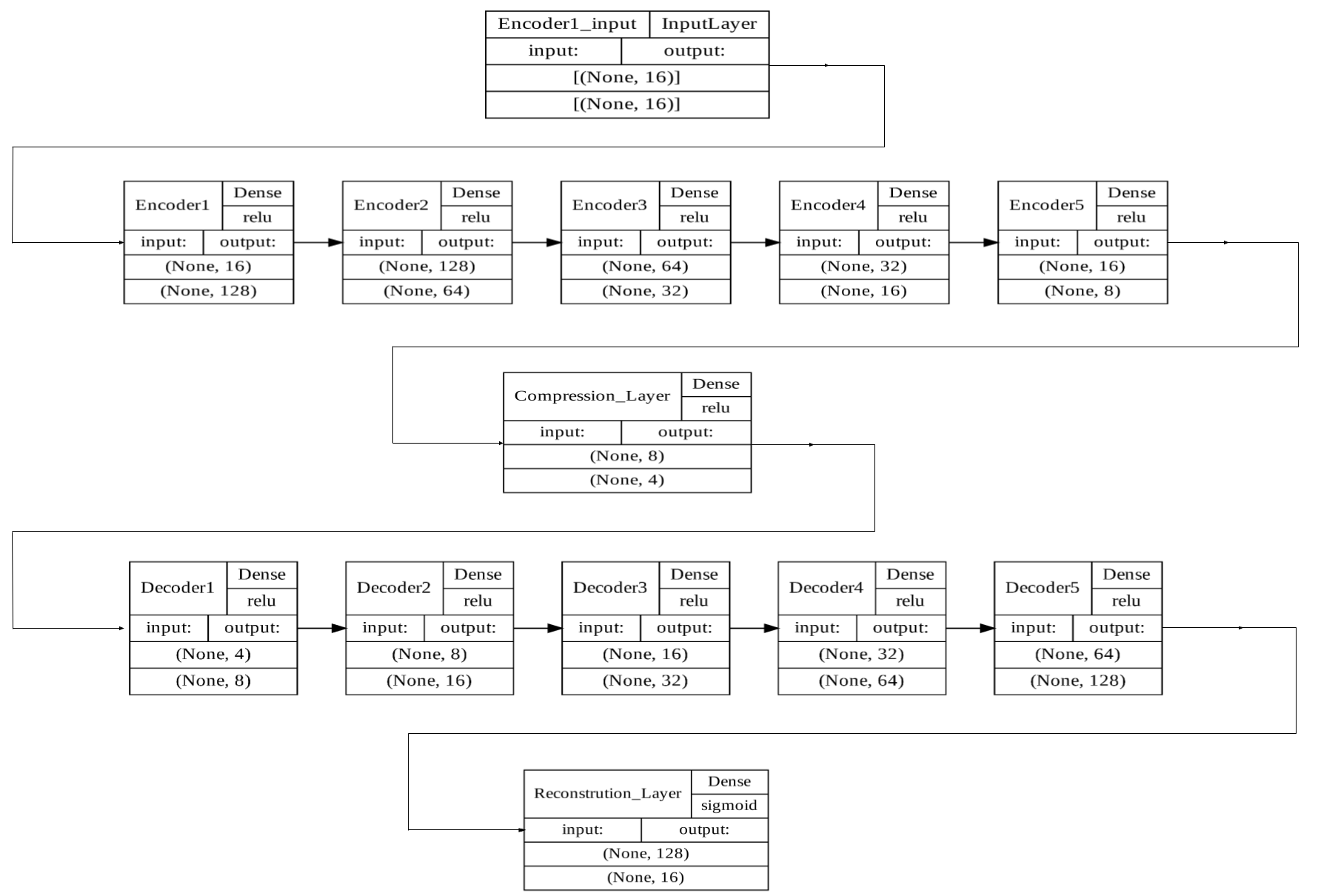}
	\caption{Architecture of the deep neural networks trained in the encoder and decoder.}
	\label{fig:pmodelau} 
\end{figure} 
\end{document}